%% file: main.tex
\documentclass[10pt,twocolumn,letterpaper]{article}

\usepackage{iccv}              


\definecolor{iccvblue}{rgb}{0.21,0.49,0.74}
\usepackage[pagebackref,breaklinks,colorlinks,allcolors=iccvblue]{hyperref}

\def\paperID{13} 
\def\confName{ICCV}
\def\confYear{2025}

\title{SIFT-Graph: Benchmarking Multimodal Defense Against Image Adversarial Attacks With Robust Feature Graph}

\author{
Jingjie He, \quad Weijie Liang, \quad Zihan Shan, \quad Matthew Caesar\\
University of Illinois Urbana-Champaign\\
{\tt\small jingjie8@illinois.edu, weijiel4@illinois.edu, zshan2@illinois.edu, caesar@illinois.edu}
}

\begin{document}
\maketitle
\input{sec/0_abstract}    
\input{sec/1_intro}

\input{sec/4_experiment}
\input{sec/5_discussion}
\input{sec/6_conclusion}
{
    \small
    \bibliographystyle{ieeenat_fullname}
    \bibliography{main}
}


\end{document}

%% file: sec/0_abstract.tex
\begin{abstract}
Adversarial attacks expose a fundamental vulnerability in modern deep vision models by exploiting their dependence on dense, pixel-level representations that are highly sensitive to imperceptible perturbations. Traditional defense strategies typically operate within this fragile pixel domain, lacking mechanisms to incorporate inherently robust visual features. In this work, we introduce SIFT-Graph, a multimodal defense framework that enhances the robustness of traditional vision models by aggregating structurally meaningful features extracted from raw images using both handcrafted and learned modalities. Specifically, we integrate Scale-Invariant Feature Transform keypoints with a Graph Attention Network to capture scale and rotation invariant local structures that are resilient to perturbations. These robust feature embeddings are then fused with traditional vision model, such as Vision Transformer and Convolutional Neural Network, to form a unified, structure-aware and perturbation defensive model. Preliminary results demonstrate that our method effectively improves the visual model robustness against gradient-based white box adversarial attacks, while incurring only a marginal drop in clean accuracy.

\end{abstract}

%% file: sec/1_intro.tex
\section{Introduction}

\label{sec:intro}
Deep learning has revolutionized computer vision, with models such as convolutional neural networks, such as ResNet~\cite{resnet}, and vision transformers ,such as ViT~\cite{vit}, achieving outstanding performance on a wide range of tasks. However, these models remain highly vulnerable to adversarial attacks~\cite{adversarial}. Carefully crafted perturbations according to model parameters, such as those generated by Projected Gradient Descent~\cite{PGD}, Fast Gradient Sign Method~\cite{FGSM}, and Auto Attack~\cite{AUTOattack}, can cause catastrophic failures in model predictions while remaining nearly imperceptible to the human eye. This vulnerability comes primarily from the models’ heavy reliance on dense, low-level pixel representations, which are inherently fragile and lack awareness of higher-level structural semantics.\\

Traditional adversarial defense strategies attempt to mitigate this vulnerability through input preprocessing~\cite{Diffusiondefense, reprocessdefense1, augmentdefense1}, gradient obfuscation~\cite{gradientobfu}, or adversarial training~\cite{PGD,adtraining1}. While achieving success in improving robustness over adversarial attack, these approaches remain constrained by the pixel domain and fail to leverage more robust visual features that are less sensitive to localized perturbations. In contrast, classical computer vision methods such as the Scale Invariant Feature Transform (SIFT)~\cite{sift} focus on extracting stable, local keypoints that are invariant to scale, rotation, and minor affine distortions. While SIFT lacks semantic richness, it captures structurally significant information that can supplement deep learned features. Building on this insight, we seek to incorporate SIFT-derived features into deep models to strengthen their resilience to adversarial perturbations.\\

We propose SIFT-Graph, a multimodal adversarial defense framework that complements conventional image-based representations with a SIFT-derived visual feature graph. We extract SIFT keypoints and construct a graph based on their spatial and descriptor, which is then processed using a Graph Attention Network (GAT)~\cite{GAT} to learn localized and resilient features to perturbation. These graph-derived representations are fused with global features from the conventional vision model (CNN and transformer) to produce a unified prediction. Our approach effectively separates scene understanding from pixel-level vulnerability by aggregating complementary information sources.\\

We evaluate SIFT-Graph on both Vision Transformers and Convolutional Neural Networks across CIFAR-10, CIFAR-100~\cite{cifar}, and Tiny ImageNet~\cite{TinyImageNet} under projected gradient descent attacks. Our method consistently enhances robustness compared to image-only modality baselines, while maintaining minimal degradation in clean accuracy. These results underscore the effectiveness of multimodal feature aggregation in defending against adversarial perturbations.

\section{Related Work}
\subsection{Adversarial Defense in Vision Models}

Despite the impressive performance of deep neural networks in image classification, they are notoriously vulnerable to adversarial examples—small, carefully crafted perturbations that can lead to incorrect predictions while remaining imperceptible to humans~\cite{PGD, adversarial}. This vulnerability poses significant risks in safety-critical applications and has motivated extensive research into defense mechanisms.

One prominent line of work is adversarial training, where models are trained on adversarial examples generated during training time~\cite{PGD}. While effective, this approach is computationally expensive and often results in a trade-off between robustness and clean accuracy. Another common method involves input preprocessing techniques such as denoising, compression, or diffusion-based restoration~\cite{Diffusiondefense}, which aim to remove adversarial noise before feeding the image into the classifier. However, many of these techniques either degrade clean image quality or are vulnerable to adaptive attacks.

A separate thread of research attempts to obfuscate gradients or modify model behavior to make it harder for attackers to compute effective perturbations~\cite{gradientobfu}. These methods have been criticized for offering ``false security," as many are easily bypassed by stronger or white-box attacks.

In addition, researchers have explored robust feature enhancement as an alternative defense strategy. Instead of relying solely on raw pixels, models can benefit from incorporating features that are less sensitive to small perturbations. Examples include frequency-domain features~\cite{freqfusion,FrequencyFusion2,FrequencyFusion}, edge maps~\cite{edgemulti1}. These approaches leverage the observation that high-frequency perturbations often fail to significantly distort structural or semantic information.

\subsection{Multimodal Learning in Computer Vision and Defense}

Multimodal learning aims to enhance model performance by integrating complementary information from multiple sources or modalities, such as vision, language, audio, and depth~\cite{multidefense1, multidefense2}. In the context of computer vision, combining modalities allows models to gain a more comprehensive understanding of the scene, which can lead to improved generalization, interpretability, and robustness.

Recent advances have leveraged joint vision-language training, as in CLIP~\cite{CLIP}, Grounding DINO~\cite{dino}, and ALBEF~\cite{ALBEF}, to learn semantically rich image embeddings aligned with textual concepts. Other efforts combine RGB and depth information~\cite{multidefense2}, or fuse spatial and frequency-domain cues~\cite{FrequencyFusion,FrequencyFusion2}, to enrich visual representations.

Multimodal approaches have also been investigated for improving robustness and adversarial defense. For instance, some works explore fusing edge information~\cite{edgemulti1}, segmentation masks~\cite{segmenteddefense}, or frequency components~\cite{freqfusion} alongside RGB images to mitigate the effects of perturbations. These studies demonstrate that adversarial examples often exploit the fragility of a single modality, and that complementary signals can compensate for compromised information pathways.

\subsection{Graph Neural Networks for Vision}

Graph Neural Networks have emerged as powerful tools for modeling relational and structured data, making them well-suited for visual tasks that involve spatial reasoning or topology-aware representations~\cite{gnnsurvey}. In computer vision, GNNs have been employed to model interactions among objects~\cite{scene_graph}, understand human-object relationships~\cite{hoi_gnn}, and perform point cloud and mesh analysis in 3D vision~\cite{3d_gnn_review}.

Unlike convolutional or transformer-based models that process grid-like image data, GNNs offer the flexibility to operate on irregular structures, such as keypoint graphs or superpixel regions. This property makes them particularly valuable for encoding geometric relationships and context beyond local receptive fields. For example, DRG-Net~\cite{siftgraph} constructs a graph over local features for diabetic retinopathy grading, while SplineCNN~\cite{splinecnn} models continuous convolution over non-Euclidean domains for shape analysis.

In adversarial settings, GNNs offer a promising defense direction due to their ability to smooth and propagate robust features across neighborhoods~\cite{robust_gnn}. Some studies have explored leveraging relational inductive bias or sparse graph structures to resist perturbations~\cite{gnn_defense}. However, few works have investigated constructing visual graphs from classical features such as SIFT, which offer a degree of invariance to scale, rotation, and local noise.

\begin{figure*}
  \centering
  \begin{subfigure}{0.68\linewidth}
    \includegraphics[width=\linewidth]{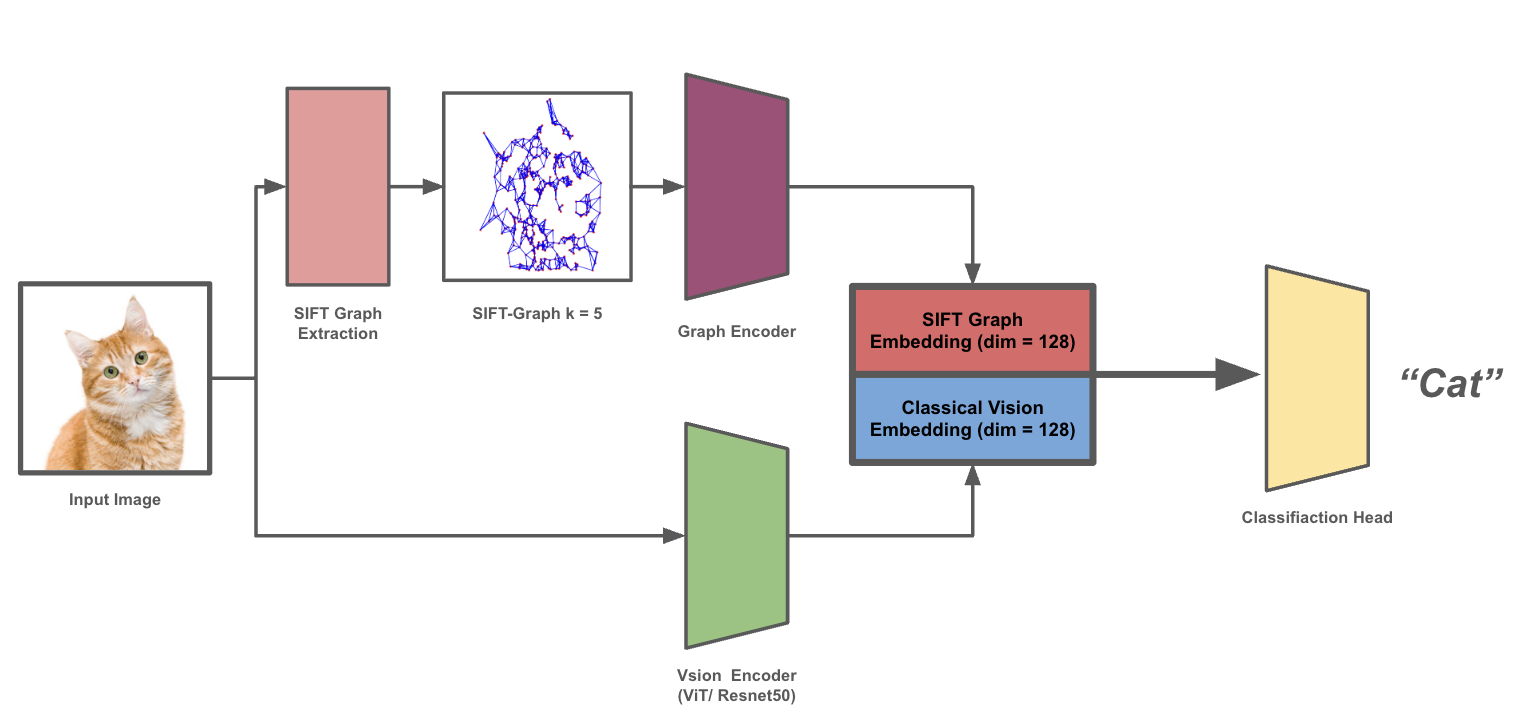}
    \caption{SIFT-Graph prediction flow}
    \label{fig:short-a}
  \end{subfigure}
  \hfill
  \begin{subfigure}{0.28\linewidth}
    \includegraphics[width=\linewidth]{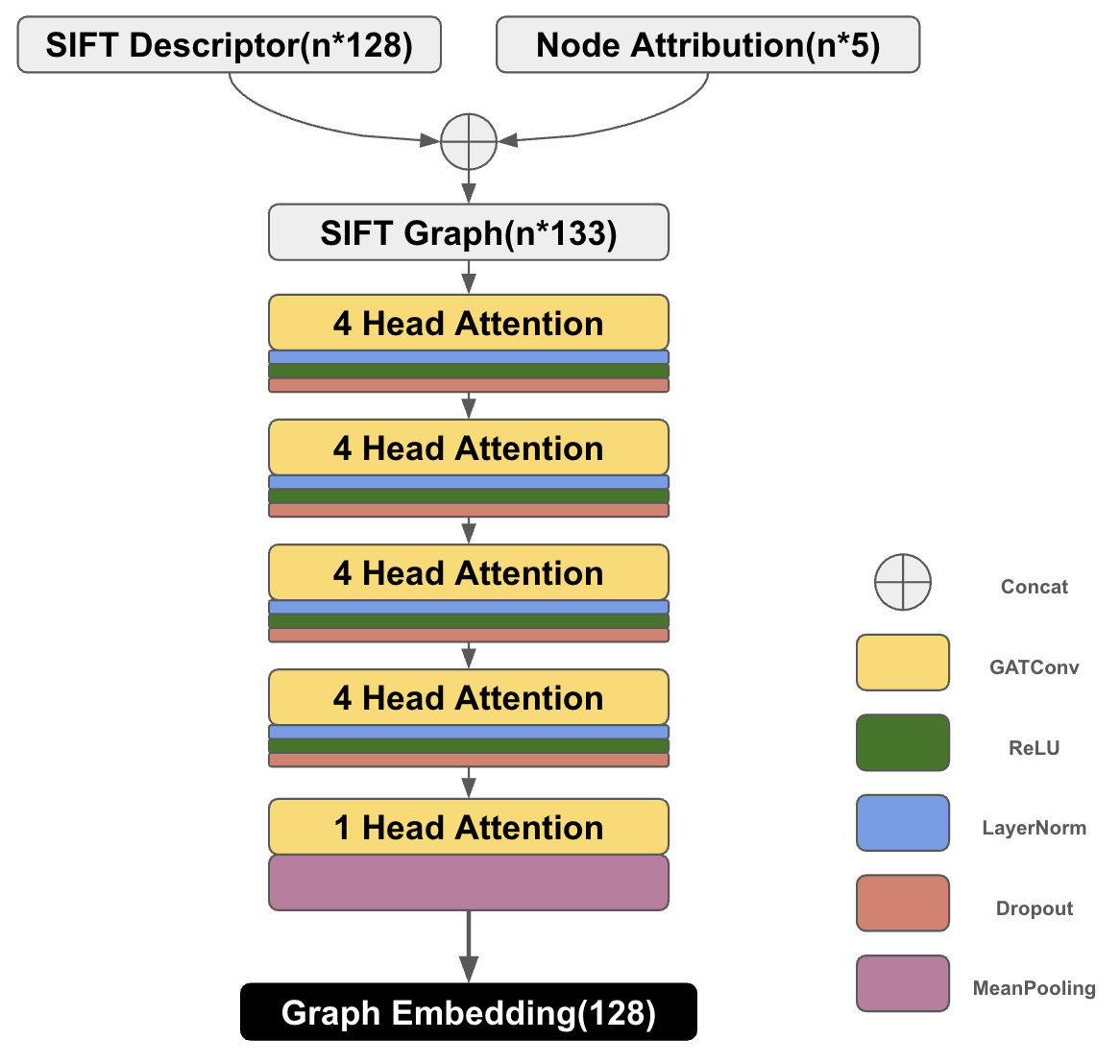}
    \caption{Graph Encoder Design}
    \label{fig:short-b}
  \end{subfigure}
  \caption{(a)Demonstration of the SIFT-Graph central model component and workflow. (b)Detailed design for graph encoder, where the node attribution refers to combination of position coordinate(2), direction(1), response(1) and size(1), where the n refers to the number of nodes.}
  \label{fig:short}
\end{figure*}

\section{Methodology}
To improve robustness against adversarial perturbations, we propose a multimodal image classification framework that combines local structural features with global semantic representations. Our method is grounded in the observation that adversarial attacks tend to exploit global shortcut cues in standard vision models, while local geometric patterns, such as those captured by SIFT keypoints, are comparatively more stable and less sensitive to pixel-level perturbations.

As illustrated in Fig.~\ref{fig:short-a}  , given an input image, we first extract SIFT keypoints and descriptors, construct a 5-nearest neighbor (k-NN) graph, and employ a GAT-based graph encoder to derive a compact and perturbation-resilient structural embedding. Concurrently, the original image is fed into a standard vision model to extract high-level semantic features. To evaluate the robustness of our multimodal system, we experiment with both ViT-B/16 and ResNet-50 as the backbone architectures for the semantic branch.

The outputs of both branches are embedded into a shared 128-dimensional space, concatenated, and passed to a classification head for final prediction. This dual-branch design allows the model to jointly reason over invariant local structures and high-level semantic context.

In the following sections, we describe each component in detail: SIFT graph construction (§3.1), graph encoder design (§3.2), semantic vision backbone (§3.3), and the fusion and classification strategy (§3.4).

\subsection{SIFT Graph Construction}
\subsubsection{SIFT Feature Extraction}

To obtain structurally meaningful features that are less sensitive to pixel-level perturbations, we extract local keypoints using the Scale-Invariant Feature Transform~\cite{sift}. Given an input image $I \in \mathbb{R}^{H \times W \times 3}$, we first convert it to grayscale $I_{\text{gray}}$, and detect a set of keypoints $\{ \mathbf{p}_i \}_{i=1}^N$, where each $\mathbf{p}_i = (x_i, y_i, s_i, \theta_i)$ represents the keypoint's spatial location, scale, and dominant orientation.

For each keypoint, SIFT computes a local descriptor $\mathbf{f}_i \in \mathbb{R}^{128}$ that encodes the gradient distribution in a $16 \times 16$ patch, subdivided into $4 \times 4$ cells with 8-bin orientation histograms:
\[
\mathbf{f}_i = \phi(I, \mathbf{p}_i) \in \mathbb{R}^{128},
\]
where $\phi(\cdot)$ denotes the SIFT descriptor operator.

In addition to the 128-dimensional descriptor $\mathbf{f}_i$, SIFT also provides spatial coordinates $(x_i, y_i)$, orientation angle $\theta_i$, response intensity $r_i$, and scale $s_i$. We also preserve these keypoint-related features to enrich the information of the graph.

\begin{figure}[t]
  \centering
  \includegraphics[width=1\linewidth]{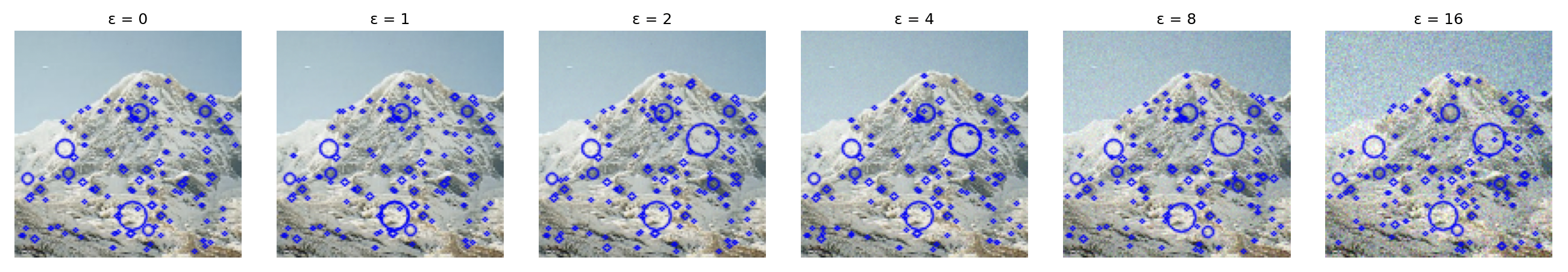}

   \caption{Example of SIFT keypoints under Gaussian noise perturbations. Each $\epsilon$ denotes the noise intensity, represented as standard deviation on the 0--255 pixel scale.
}
   \label{fig:onecol}
\end{figure}

\paragraph{Stability Under Perturbation.}
SIFT descriptors are designed to be invariant to image scaling, rotation, and minor affine transformations. More importantly for adversarial settings, they are less sensitive to small, structured perturbations~\cite{siftrobust} than raw pixel values, as also illustrated in Fig.~\ref{fig:onecol}.  

Let $I' = I + \delta$ denote a perturbed image, where $\| \delta \|_\infty \leq \epsilon$. While deep models operating directly on pixels often exhibit large changes in activation from such perturbations, SIFT descriptors are relatively stable:
\[
\| \phi(I, \mathbf{p}_i) - \phi(I', \mathbf{p}_i') \|_2 \ll \| I - I' \|_2,
\]
as long as the perturbed keypoint $\mathbf{p}_i'$ remains close to the original $\mathbf{p}_i$ in space. In practice, we observe that keypoints may slightly shift, but continue to lie on salient structures such as corners and edges, preserving relational geometry for subsequent graph construction.

This structural consistency motivates the use of SIFT keypoints as robust anchors for graph-based feature aggregation under adversarial conditions.
\subsubsection{Graph Construction}

Given a set of $N$ SIFT keypoints and their associated feature vectors $\{ \mathbf{x}_i \in \mathbb{R}^{133} \}_{i=1}^N$, we construct a $k$-nearest neighbor (k-NN) \cite{knn} graph  $\mathcal{G} = (\mathcal{V}, \mathcal{E})$, where each node $v_i \in \mathcal{V}$ corresponds to a keypoint.

\begin{figure*}[t]
  \centering
   \includegraphics[width=1\linewidth]{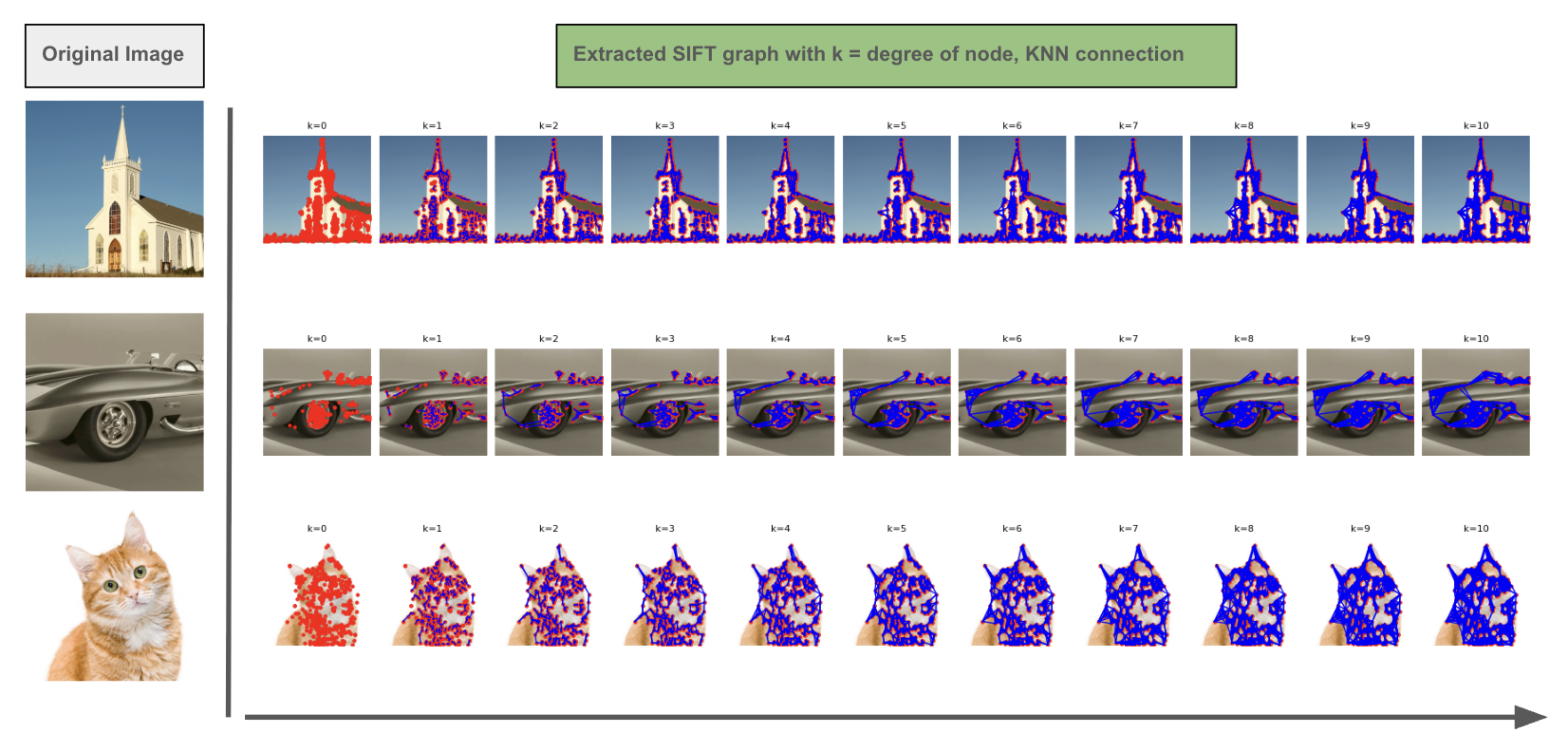}

   \caption{Visualization of SIFT-based $k$-nearest neighbor graphs constructed with varying values of $k$. }
   \label{fig:short}
\end{figure*}

\paragraph{Node Features} 
Each node $v_i$ is associated with a feature vector $\mathbf{x}_i \in \mathbb{R}^{133}$, which includes the 128-dimensional SIFT descriptor $\mathbf{f}_i$, normalized 2D coordinates, and additional local attributes: orientation angle, response intensity, and keypoint scale. Formally, the node feature is defined as:
\[
\mathbf{x}_i = \begin{bmatrix}
\mathbf{f}_i \\
\hat{x}_i \\
\hat{y}_i \\
\theta_i \\
r_i \\
s_i
\end{bmatrix}, \quad 
\text{with} \quad 
\hat{x}_i = \frac{x_i - \mu_x}{\sigma_x + \epsilon}, \quad 
\hat{y}_i = \frac{y_i - \mu_y}{\sigma_y + \epsilon},
\]
where $(x_i, y_i)$ are the raw keypoint coordinates, $\theta_i$ is the orientation angle, $r_i$ is the response intensity, and $s_i$ is the keypoint scale. The coordinate normalization is computed using the dataset-wide mean $(\mu_x, \mu_y)$ and standard deviation $(\sigma_x, \sigma_y)$. A small constant $\epsilon$ is added to avoid division by zero.

\paragraph{Edge Construction:}
To define the edge set $\mathcal{E}$, we apply k-nearest neighbor search in the normalized spatial domain. We avoid using fully connected graphs, as the computational cost scales quadratically with the number of nodes. For large graphs, this results in significant overhead that diminishes the practical efficiency and scalability of the approach. An undirected edge $(i, j)$ is added between nodes $v_i$ and $v_j$ if $v_j$ is among the $k$ nearest neighbors of $v_i$:
\[
(i, j) \in \mathcal{E} \quad \Leftrightarrow \quad v_j \in \text{Top-}k \left( \text{dist}(\hat{\mathbf{p}}_i, \hat{\mathbf{p}}_j) \right),
\]
where $\hat{\mathbf{p}}_i = (\hat{x}_i, \hat{y}_i)$ are the normalized coordinates and $\text{dist}(\cdot)$ is the Euclidean distance in 2D space. In our implementation, we symmetrize the graph to ensure undirected connectivity:
\[
(i, j) \in \mathcal{E} \quad \text{if} \quad v_j \in \mathcal{N}_k(v_i) \quad \text{or} \quad v_i \in \mathcal{N}_k(v_j).
\]
To select an appropriate value for k, we empirically set k=5, as shown in Fig.~\ref{fig:short} . According to Geiringer–Laman theorem~\cite{GLaman1970}, in a 2D space, at least three edges are required to anchor the relative spatial configuration of a node. In practice, setting  k=5 ensures a sufficiently dense graph structure that preserves local neighborhood information while avoiding excessive redundancy or overly repetitive edge connections.
\paragraph{Graph Representation:}
The resulting graph $\mathcal{G}$ is represented as a tuple $(\mathbf{X}, \mathbf{E})$, where $\mathbf{X} \in \mathbb{R}^{N \times 133}$ contains the node features, and $\mathbf{E} \in \mathbb{N}^{2 \times |\mathcal{E}|}$ is the edge index tensor used by graph neural network backends.

This construction captures local spatial relationships among robust keypoints and facilitates neighborhood-aware aggregation in the subsequent GAT module.
\subsection{Graph Encoder Design}

After constructing the SIFT-based keypoint graph $\mathcal{G} = (\mathcal{V}, \mathcal{E})$, we employ a Graph Attention Network \cite{GAT} to encode robust structural embeddings through attention-weighted message passing. As shown in Fig.~\ref{fig:short-b}  , we employ a stack of five GATConv layers, each incorporating a multi-head attention mechanism. This design enables the model to effectively capture local and global structural dependencies within the graph. Each GATConv layer computes attention-weighted feature propagation from neighboring nodes, allowing for dynamic, data-driven aggregation. The output of each layer is passed through non-linear activation and normalization, facilitating stable and expressive graph-level feature learning. This layered configuration supports progressive integration of node-level information into a coherent graph-level representation.

\paragraph{Graph Attention Mechanism:}
Let $\mathbf{h}_i^{(0)} = \mathbf{x}_i$ denote the initial feature of node $v_i$, combining the SIFT descriptor and normalized spatial coordinates. A GAT layer updates node features according to:
\[
\mathbf{h}_i^{(l+1)} = \sigma \left( \sum_{j \in \mathcal{N}(i)} \alpha_{ij}^{(l)} \mathbf{W}^{(l)} \mathbf{h}_j^{(l)} \right),
\]
where $\mathcal{N}(i)$ denotes the set of neighboring nodes of $v_i$, $\mathbf{W}^{(l)}$ is a learnable weight matrix for the $l$-th layer, $\sigma(\cdot)$ is a non-linear activation function such as ReLU, and $\alpha_{ij}^{(l)}$ is the normalized attention coefficient that determines the relative importance of node $j$'s features to node $i$ in the $l$-th layer.

The attention coefficient is computed via a shared feed-forward network:
\[
e_{ij}^{(l)} = \text{LeakyReLU} \left( \mathbf{a}^\top \left[ \mathbf{W}^{(l)} \mathbf{h}_i^{(l)} \, \| \, \mathbf{W}^{(l)} \mathbf{h}_j^{(l)} \right] \right),
\]
\[
\alpha_{ij}^{(l)} = \frac{\exp(e_{ij}^{(l)})}{\sum_{k \in \mathcal{N}(i)} \exp(e_{ik}^{(l)})},
\]
where $\|$ denotes vector concatenation and $\mathbf{a}$ is a learnable vector parameter.

\paragraph{Multi-Head Aggregation:}
We use multi-head attention to stabilize learning and capture diverse subspace interactions:
\[
\mathbf{h}_i^{(l+1)} = \Big\|_{m=1}^{M} \sigma \left( \sum_{j \in \mathcal{N}(i)} \alpha_{ij}^{(l, m)} \mathbf{W}^{(l, m)} \mathbf{h}_j^{(l)} \right),
\]
where $M$ is the number of attention heads, each with its own parameters.

\paragraph{Global Graph Embedding:}
After $L$ layers of attention-based aggregation, we obtain node embeddings $\mathbf{h}_i^{(L)}$ for all $v_i \in \mathcal{V}$. A global graph representation is computed using mean or max pooling over all nodes:
\[
\mathbf{z}_{\text{GAT}} = \text{Pooling} \left( \{ \mathbf{h}_i^{(L)} \}_{i=1}^N \right).
\]

This representation captures robust structural information from spatially grounded keypoints, enabling effective defense against perturbations that disrupt raw pixel correlations but preserve local structure.

\subsection{Semantic Vision Backbone}

To capture global semantic information from the input image, we adopt standard deep vision models as the semantic branch of our framework. Specifically, we experiment with both CNN and ViT to evaluate the generalizability of our method across different backbone architectures.

For the CNN-based branch, we use ResNet-50~\cite{resnet} as a representative architecture. The input image $I \in \mathbb{R}^{H \times W \times 3}$ is passed through a series of convolutional blocks and residual layers to produce a high-level feature map. We apply global average pooling to obtain a fixed-dimensional vector $\mathbf{z}_{\text{CNN}} \in \mathbb{R}^{d}$ as the semantic embedding.

For the transformer-based branch, we adopt ViT-B/16~\cite{vit}, a standard vision transformer that processes the image as a sequence of non-overlapping patches. The image is divided into $N$ patches of size $P \times P$, each linearly projected into an embedding vector. Positional encodings are added, and a special [CLS] token is prepended to the sequence:
\[
\mathbf{z}_0 = [\mathbf{e}_{\text{[CLS]}} ; \mathbf{E} \mathbf{p}_1 + \mathbf{e}_1 ; \ldots ; \mathbf{E} \mathbf{p}_N + \mathbf{e}_N],
\]
where $\mathbf{p}_i$ is the $i$-th patch, $\mathbf{E}$ is the shared projection matrix, and $\mathbf{e}_i$ are the positional encodings. The sequence is processed by $L$ transformer layers composed of multi-head self-attention and feedforward submodules. The final embedding of the [CLS] token is taken as the semantic representation:
\[
\mathbf{z}_{\text{ViT}} = \mathbf{z}^{(L)}_{\text{[CLS]}} \in \mathbb{R}^{d}.
\]

In both cases, the semantic branch produces a feature vector $\mathbf{z}_{\text{sem}} \in \mathbb{R}^{d}$(either $\mathbf{z}_{\text{CNN}}$ or $\mathbf{z}_{\text{ViT}}$) that encodes global scene understanding, where we set $d = 128$ to match the dimensionality of the graph embedding. This dimensionality is deliberately aligned with the output of the graph encoder to ensure balanced contribution from both modalities, allowing the semantic embedding to be effectively fused with the structural graph representation for robust classification.

\subsection{Embedding Fusion and Classification}

To produce a final prediction, we fuse the semantic and structural representations obtained from the vision backbone and the graph encoder, respectively. Let $\mathbf{z}_{\text{sem}} \in \mathbb{R}^{d}$ denote the semantic embedding (from either ResNet-50 or ViT-B/16), and let $\mathbf{z}_{\text{GAT}} \in \mathbb{R}^{d}$ represent the graph embedding obtained via the GAT-based encoder. Both embeddings are projected into a shared feature space of dimension $d = 128$, ensuring equal contribution from both modalities.

We concatenate the two representations to form a unified multimodal feature vector:
\[
\mathbf{z}_{\text{fused}} = [\mathbf{z}_{\text{sem}} \, \| \, \mathbf{z}_{\text{GAT}}] \in \mathbb{R}^{2d}.
\]

This fused vector is then passed through a multilayer perceptron classifier consisting of a linear projection, followed by a non-linear activation and a final output layer:
\[
\hat{\mathbf{y}} = \text{MLP}(\mathbf{z}_{\text{fused}}) = \mathbf{W}_2 \cdot \sigma(\mathbf{W}_1 \cdot \mathbf{z}_{\text{fused}} + \mathbf{b}_1) + \mathbf{b}_2,
\]
where $\mathbf{W}_1 \in \mathbb{R}^{h \times 2d}$ and $\mathbf{W}_2 \in \mathbb{R}^{C \times h}$ are learnable weights, $\mathbf{b}_1$ and $\mathbf{b}_2$ are biases, $\sigma(\cdot)$ is a non-linear activation function (e.g., ReLU), $h$ is the hidden dimension, and $C$ is the number of output classes.

The model is trained end-to-end using the cross-entropy loss between the predicted logits $\hat{\mathbf{y}}$ and the ground-truth label $\mathbf{y}$. During training, gradients are backpropagated through both the vision backbone and the graph encoder, enabling joint optimization of both branches for improved adversarial robustness and clean accuracy.

%% file: sec/4_experiment.tex
\section{Experiment and Results}

\subsection{Experimental Setup}

To evaluate the effectiveness of the proposed SIFT-Graph framework, we selected two commonly used backbone architectures: ViT-B/16~\cite{vit} and ResNet-50~\cite{resnet}. For each dataset, CIFAR-10~\cite{cifar}, CIFAR-100~\cite{cifar}, and Tiny ImageNet~\cite{TinyImageNet}, we used pretrained model checkpoints that achieve highest clean accuracy as our baselines.

After integrating SIFT-Graph into each backbone, we performed lightweight fine-tuning on the corresponding dataset. This step allows the modified models to adapt to the additional structural components and multimodal inputs introduced by SIFT-Graph, while keeping the overall architecture and training procedure mostly unchanged.

We evaluated both the original and the SIFT-Graph-enhanced models on clean and adversarial examples. For evaluation of adversarial robustness, we adopt a white-box threat model, where the attacker has full access to the model parameters and gradients. Specifically, we use the projected gradient descent (PGD) attack~\cite{PGD}, a strong first-order iterative attack. To assess model robustness under varying levels of perturbation, we vary the maximum perturbation budget $\epsilon$ over seven values sampled logarithmically from 0.001 to 0.1.

All evaluations are conducted under the same PGD configuration to ensure consistent and fair comparison among different variants of the model.

\subsection{Results}

We evaluate the impact of SIFT-Graph on model robustness and clean accuracy across three datasets (CIFAR-10, CIFAR-100, and Tiny ImageNet) and two backbone architectures (ViT-B/16 and ResNet-50). We report the performance under PGD attacks with increasing perturbation strength $\epsilon$ in Tables~\ref{tab:vit_results} and~\ref{tab:res_results}, and visualize the corresponding accuracy trends in Figures~\ref{fig:vit-all} and~\ref{fig:res-all}.

\begin{table*}[ht]
\centering
\caption{Accuracy under PGD Attack for ViT and SIFT+ViT models across different $\epsilon$ values. Best results per setting are bolded.}
\label{tab:vit_results} 
\begin{tabular}{l|cc|cc|cc}
\toprule
Attack Radius $\epsilon$ & \multicolumn{2}{c|}{CIFAR-10} & \multicolumn{2}{c|}{CIFAR-100} & \multicolumn{2}{c}{Tiny ImageNet} \\
                         & SIFT+ViT & ViT & SIFT+ViT & ViT & SIFT+ViT & ViT \\
\midrule
$\epsilon = 0.0000$ & 0.9679 & \textbf{0.9788} & 0.8471 & \textbf{0.8985} & 0.8053 & \textbf{0.8471} \\
$\epsilon = 0.0010$ & \textbf{0.6064} & 0.4840 & 0.2932 & \textbf{0.3470} & \textbf{0.2310} & 0.1650 \\
$\epsilon = 0.0018$ & \textbf{0.5737} & 0.3360 & 0.2639 & \textbf{0.2825} & \textbf{0.2124} & 0.1585 \\
$\epsilon = 0.0032$ & \textbf{0.5139} & 0.1560 & \textbf{0.2197} & 0.1920 & \textbf{0.1870} & 0.1068 \\
$\epsilon = 0.0056$ & \textbf{0.4106} & 0.0410 & \textbf{0.1519} & 0.0922 & \textbf{0.1423} & 0.0824 \\
$\epsilon = 0.0100$ & \textbf{0.2458} & 0.0080 & \textbf{0.0881} & 0.0272 & \textbf{0.0896} & 0.0120 \\
$\epsilon = 0.0178$ & \textbf{0.1086} & 0.0000 & \textbf{0.0417} & 0.0033 & \textbf{0.0437} & 0.0060 \\
$\epsilon = 0.0316$ & \textbf{0.0278} & 0.0000 & \textbf{0.0154} & 0.0007 & \textbf{0.0176} & 0.0010 \\
$\epsilon = 0.0562$ & \textbf{0.0054} & 0.0000 & \textbf{0.0046} & 0.0005 & \textbf{0.0037} & 0.0000 \\
$\epsilon = 0.1000$ & \textbf{0.0007} & 0.0000 & \textbf{0.0015} & 0.0003 & 0.0000 & 0.0000 \\
\bottomrule
\end{tabular}
\end{table*}

\subsubsection{SIFTGraph on Vision Transformer}
\begin{figure*}[t]
  \centering

  \begin{subfigure}[t]{0.32\linewidth}
    \includegraphics[width=\linewidth]{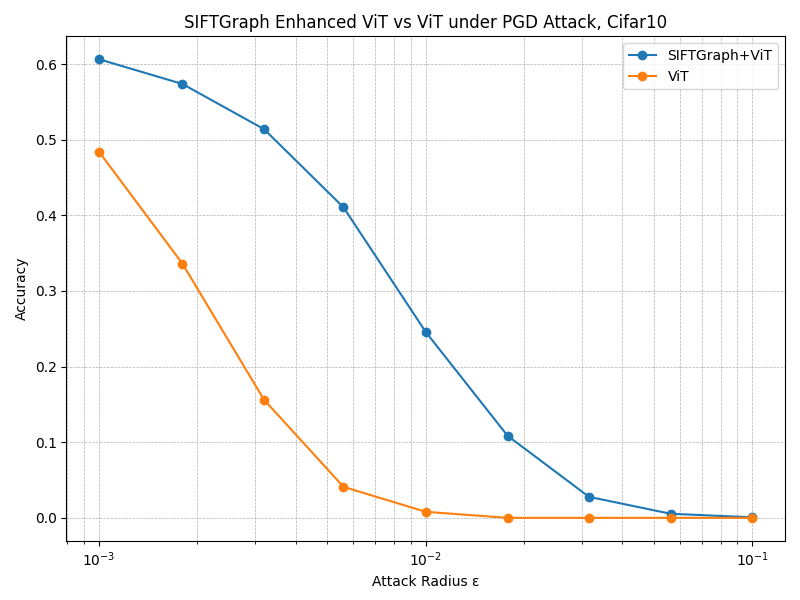}
    \caption{CIFAR-10}
    \label{fig:subfig1}
  \end{subfigure}
  \hfill
  \begin{subfigure}[t]{0.32\linewidth}
    \includegraphics[width=\linewidth]{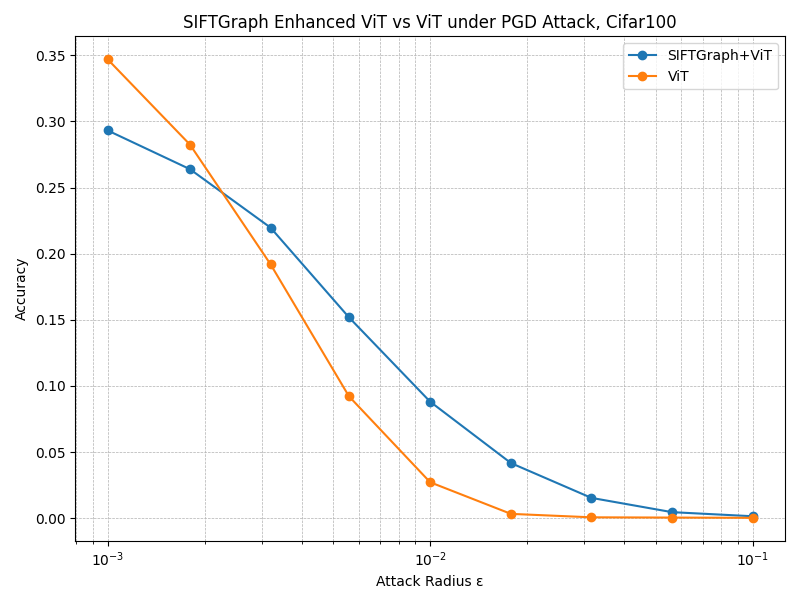}
    \caption{CIFAR-100}
    \label{fig:subfig2}
  \end{subfigure}
  \hfill
  \begin{subfigure}[t]{0.32\linewidth}
    \includegraphics[width=\linewidth]{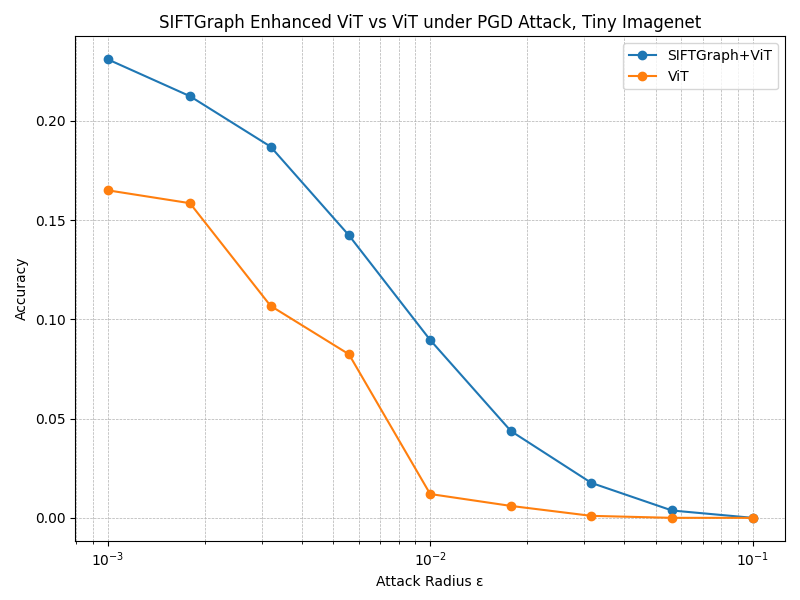}
    \caption{Tiny ImageNet}
    \label{fig:subfig3}
  \end{subfigure}

  \caption{Robustness evaluation under PGD attacks for ViT vs. SIFTGraph enhanced ViT on multiple datasets.}
  \label{fig:vit-all}
\end{figure*}

For ViT-B/16, incorporating SIFT-Graph consistently improves adversarial robustness across all datasets and perturbation levels. Although there is a slight decrease in clean accuracy (e.g., from 97.88\% to 96.79\% on CIFAR-10), the resilience under attack is substantially enhanced. Notably, at $\epsilon=0.0056$, the CIFAR-10 accuracy improves from 4.1\% to 41.06\%, a tenfold increase. Similar improvements are observed on CIFAR-100 and Tiny ImageNet (Table~\ref{tab:vit_results}, Figure~\ref{fig:vit-all}).

This significant gain suggests that local structural features captured by SIFT-Graph complement the global attention mechanism of transformers, enhancing their ability to resist adversarial perturbations.

\subsubsection{SIFTGraph on Convolutional Neural Network}
\begin{table*}[ht]
\centering
\caption{Accuracy under PGD Attack for ResNet and SIFT+ResNet models across different $\epsilon$ values. Best results per setting are bolded.}
\label{tab:res_results} 
\begin{tabular}{l|cc|cc|cc}
\toprule
Attack Radius $\epsilon$ & \multicolumn{2}{c|}{CIFAR-10} & \multicolumn{2}{c|}{CIFAR-100} & \multicolumn{2}{c}{Tiny ImageNet} \\
                         & SIFT+ResNet & ResNet & SIFT+ResNet & ResNet & SIFT+ResNet & ResNet \\
\midrule
$\epsilon = 0.0000$ & \textbf{0.9587} & 0.9465 & 0.7909 & \textbf{0.8093} & \textbf{0.6854} & 0.6580 \\
$\epsilon = 0.0010$ & \textbf{0.4019} & 0.1706 & \textbf{0.0972} & 0.0703 & \textbf{0.0566} & 0.0235 \\
$\epsilon = 0.0018$ & \textbf{0.3181} & 0.1056 & \textbf{0.0708} & 0.0492 & \textbf{0.0398} & 0.0195 \\
$\epsilon = 0.0032$ & \textbf{0.1997} & 0.0655 & \textbf{0.0461} & 0.0266 & \textbf{0.0261} & 0.0168 \\
$\epsilon = 0.0056$ & \textbf{0.0979} & 0.0125 & \textbf{0.0242} & 0.0164 & \textbf{0.0146} & 0.0120 \\
$\epsilon = 0.0100$ & \textbf{0.0371} & 0.0070 & \textbf{0.0127} & 0.0109 & \textbf{0.0056} & 0.0020 \\
$\epsilon = 0.0178$ & \textbf{0.0134} & 0.0047 & \textbf{0.0068} & 0.0070 & \textbf{0.0029} & 0.0000 \\
$\epsilon = 0.0316$ & \textbf{0.0068} & 0.0023 & \textbf{0.0042} & 0.0023 & \textbf{0.0005} & 0.0000 \\
$\epsilon = 0.0562$ & \textbf{0.0037} & 0.0016 & \textbf{0.0027} & 0.0016 & \textbf{0.0005} & 0.0000 \\
$\epsilon = 0.1000$ & \textbf{0.0020} & 0.0000 & \textbf{0.0007} & 0.0000 & 0.0000 & 0.0000 \\
\bottomrule
\end{tabular}
\end{table*}

\begin{figure*}[t]
  \centering

  \begin{subfigure}[t]{0.33\linewidth}
    \includegraphics[width=\linewidth]{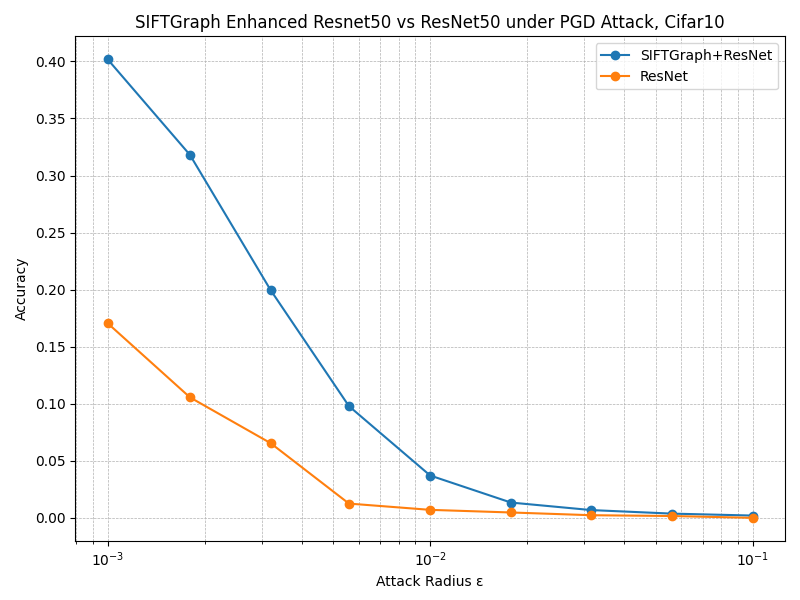}
    \caption{CIFAR-10}
    \label{fig:subfig1}
  \end{subfigure}
  \hfill
  \begin{subfigure}[t]{0.33\linewidth}
    \includegraphics[width=\linewidth]{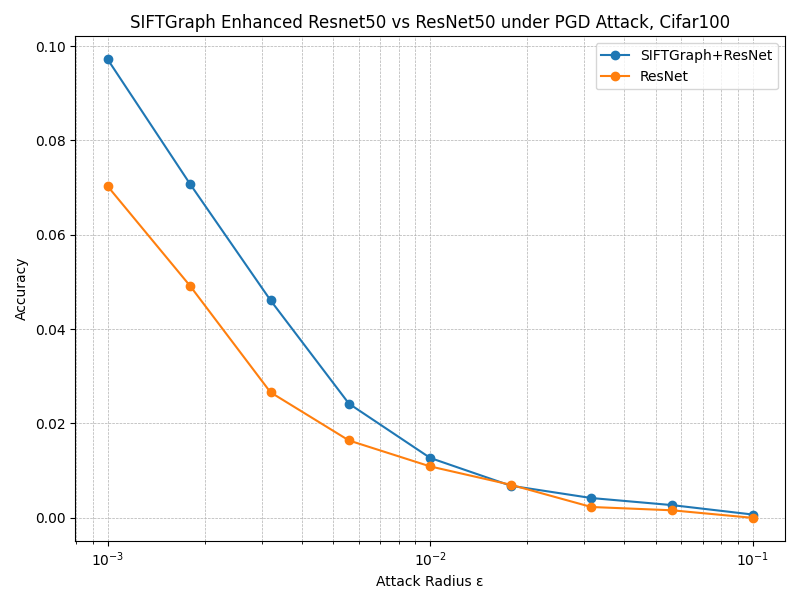}
    \caption{CIFAR-100}
    \label{fig:subfig2}
  \end{subfigure}
  \hfill
  \begin{subfigure}[t]{0.33\linewidth}
    \includegraphics[width=\linewidth]{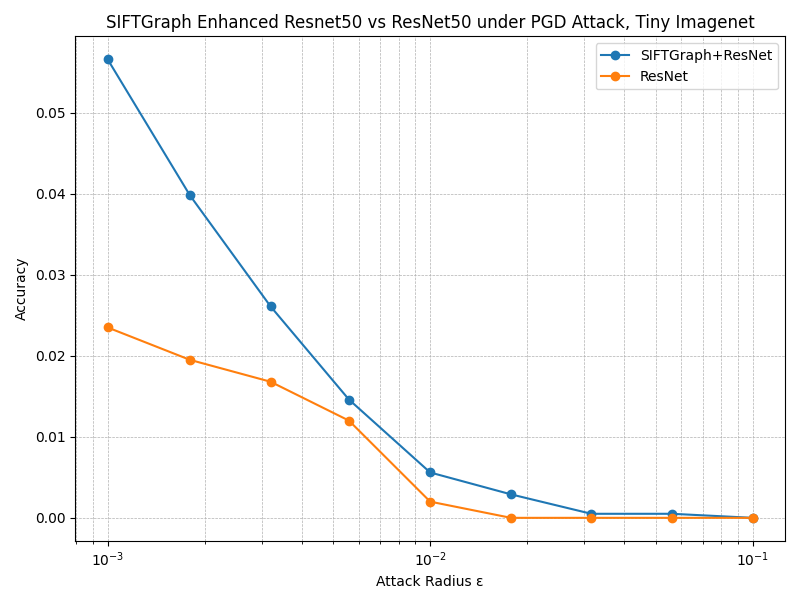}
    \caption{Tiny ImageNet}
    \label{fig:res-all}
  \end{subfigure}

  \caption{Robustness evaluation under PGD attacks for ResNet50 vs. SIFTGraph enhanced ResNet50 on multiple datasets.}
  \label{fig:res-all}
\end{figure*}

ResNet-50 models augmented with SIFT-Graph show similar trends. Clean accuracy is preserved or slightly improved, such as on CIFAR-10 where accuracy increases from 94.65\% to 95.87\%. Under PGD attack, SIFT+ResNet outperforms the baseline across all tested $\epsilon$ values. At $\epsilon=0.0032$, for example, accuracy on CIFAR-10 increases from 6.55\% to 19.97\% (Table~\ref{tab:res_results}, Figure~\ref{fig:res-all}).

Although the magnitude of robustness improvement is slightly smaller than that seen with ViT, the consistent performance gain highlights the value of explicit local feature integration even in architectures that already encode locality through convolution.

\subsection{Summary}

Our experiments demonstrate that SIFT-Graph is a lightweight and effective addition to vision models, yielding substantial gains in adversarial robustness without requiring adversarial training or architectural changes. The performance improvements are consistent across datasets and backbones, with particularly pronounced gains on transformer-based models like ViT-B/16.

By leveraging robust local keypoints and structural cues, SIFT-Graph enhances the model's capacity to resist perturbations while maintaining competitive clean accuracy. These results underscore the value of combining handcrafted local features with modern deep architectures for multimodal adversarial defense.

%% file: sec/5_discussion.tex
\section{Discussion}
Our experiments highlight both the strengths and limitations of the proposed SIFT-Graph framework.One clear advantage is its lightweight and efficient design. SIFT-Graph is an independent plug-and-play module that can be seamlessly integrated into multiple mainstream architectures such as ViT-B/16 and ResNet-50 without introducing additional depth or structural complexity. This makes it easy to adopt in practice and compatible with a wide range of backbone models. Moreover, even with only modest fine-tuning, meaning that the module's full potential may not yet be fully realized, SIFT-Graph already yields significant robustness improvements under PGD attacks \cite{PGD}. This suggests that more extensive training or task-specific tuning could further enhance its effectiveness. In the future, we also plan to explore the application of SIFT-Graph to other vision tasks or the integration of it with other popular backbone architectures beyond classification.

Another strength lies in its minimal impact on standard performance. Unlike many advanced defense methods (e.g., Diffusiondefense \cite{Diffusiondefense}) that achieve high robustness at the cost of reduced clean accuracy, SIFT-Graph maintains comparable or even slightly improved performance on clean data. This property is crucial for practical deployment, as it ensures the model remains reliable under standard conditions while gaining adversarial resilience.

However, there are important limitations to acknowledge. First, the SIFT preprocessing itself introduces significant information loss by reducing rich pixel-level input to sparse keypoints and descriptors. While this simplification helps block certain adversarial perturbations, it inherently limits the potential performance gains. Second, while SIFT-Graph is highly effective against attacks like PGD \cite{PGD} that apply small, imperceptible perturbations, it performs less well against adversarial strategies that visibly alter image structure and texture. In these scenarios, the handcrafted local features from SIFT may not capture enough reliable information to ensure robust classification.

Overall, while SIFT-Graph offers a promising, efficient, and easily integrable approach to improving adversarial robustness, future work will need to address its information bottleneck and evaluate its effectiveness against a broader range of attack types.

%% file: sec/6_conclusion.tex
\section{Conclusion}
In this work, we introduced SIFT-Graph, a simple and effective framework for enhancing the adversarial robustness of vision models through multimodal feature integration. By incorporating handcrafted local features into deep network backbones, SIFT-Graph improves resilience to adversarial perturbations without requiring major architectural changes or adversarial training. Our experiments on three datasets demonstrated consistent gains in robustness across both transformer-based and convolutional architectures, with minimal impact on clean accuracy.

While SIFT-Graph offers a lightweight and easily deployable defense strategy, our results also highlight its limitations, such as potential information loss during SIFT extraction and reduced effectiveness against attacks that heavily distort image structure. Future work will explore improving the feature extraction pipeline, adapting SIFT-Graph to more challenging attack scenarios, and extending it to a broader range of vision tasks and architectures.

Overall, SIFT-Graph represents a promising direction for leveraging classical local features to complement modern deep learning models, providing a practical step toward more robust and interpretable adversarial defenses.